\documentclass[sigconf]{acmart}

\usepackage{hyperref}
\usepackage{url}
\usepackage{graphicx}
\usepackage{amsthm}
\usepackage{algorithm}
\usepackage{algorithmic}
\usepackage{caption}
\usepackage{subcaption}
\usepackage{wrapfig}

\usepackage{amsmath,amsfonts,bm}

\def\eqref#1{equation~\ref{#1}}

\def\1{\bm{1}}

\def\mX{{\bm{X}}}
\def\mY{{\bm{Y}}}

\DeclareMathAlphabet{\mathsfit}{\encodingdefault}{\sfdefault}{m}{sl}
\SetMathAlphabet{\mathsfit}{bold}{\encodingdefault}{\sfdefault}{bx}{n}

\def\gE{{\mathcal{E}}}

\def\gG{{\mathcal{G}}}

\def\gI{{\mathcal{I}}}

\def\gR{{\mathcal{R}}}

\def\gV{{\mathcal{V}}}

\def\gX{{\mathcal{X}}}
\def\gY{{\mathcal{Y}}}

\def\sI{{\mathbb{I}}}

\def\sP{{\mathbb{P}}}

\def\sR{{\mathbb{R}}}

\newcommand{\E}{\mathbb{E}}

\newcommand{\R}{\mathbb{R}}

\newtheorem{definition}{Definition}

\newcommand{\argtopk}{\mathop{\mathrm{arg\,top}\,k}}

\newcommand{\sss}{Task-aware CL}
\newcommand{\rtree}{RelStump}
\newcommand{\xgs}{XGSampler}
\newcommand{\loss}{XTCL}

\newcommand{\jn}[1]{}
\renewcommand{\jn}[1]{{\color{red} JN: {#1}}}

\newcommand{\jean}[1]{}
\renewcommand{\jean}[1]{{\color{green} JL: {#1}}}

\newcommand{\first}[1]{}
\renewcommand{\first}[1]{{\textbf{\color{blue}{#1}}}}

\newcommand{\second}[1]{}
\renewcommand{\second}[1]{{\color{blue}{#1}}}

\newcommand{\third}[1]{}
\renewcommand{\third}[1]{{\color{violet}{#1}}}

\AtBeginDocument{%
  }

\setcopyright{acmlicensed}
\copyrightyear{2018}
\acmYear{2018}
\acmDOI{XXXXXXX.XXXXXXX}

\acmConference[under review]{Make sure to enter the correct
  conference title from your rights confirmation emai}{October,2024}{WWW}

\begin{document}

\title{Improving Node Representation by Boosting Task-Aware Contrastive Loss}


\author{Ying-Chun Lin}
\affiliation{%
  \institution{Purdue University}
  \city{West Lafayette}
  \country{USA}}
\email{lin915@purdue.edu}

\author{Jennifer Neville}
\affiliation{%
  \institution{Microsoft Research}
  \city{Redmond}
  \country{USA}}
\email{jenneville@microsoft.com}

\renewcommand{\shortauthors}{Lin et al.}

\begin{abstract}
Graph datasets record complex relationships between entities, with nodes and edges capturing intricate associations. Node representation learning involves transforming nodes into low-dimensional embeddings. These embeddings are typically used as features for downstream tasks. Therefore, their quality has a significant impact on task performance. Existing approaches for node representation learning span (semi-)supervised, unsupervised, and self-supervised paradigms. In graph domains, (semi-)supervised learning typically optimizes models based on class labels, neglecting other abundant graph signals, which limits generalization. While self-supervised or unsupervised learning produces representations that better capture underlying graph signals, the usefulness of these captured signals for downstream tasks can vary. To bridge this gap, we introduce Task-Aware Contrastive Learning (\sss), which aims to enhance downstream task performance by maximizing the mutual information between the downstream task and node representations with a self-supervised learning process. This is achieved through a sampling function, XGBoost Sampler (\xgs), to sample proper positive examples for the proposed Task-Aware Contrastive Loss (\loss). By minimizing {\loss}, {\sss} increases the mutual information between the downstream task and node representations and captures useful graph signals into the node representations, such that model generalization is improved. Additionally, {\xgs} enhances the interpretability of each signal by showing the weights for sampling the proper positive examples. We show experimentally that {\loss} significantly improves performance on two downstream tasks: node classification and link prediction tasks, compared to state-of-the-art models. Our code is available on GitHub\footnote{Github link will be provided after publication to preserve anonymity.}.
\end{abstract}



\keywords{Graph Neural Network, Contrastive Learning}



\maketitle

\section{Introduction}

A graph is a data structure that represents relationships between entities or elements using nodes and edges. Node representation learning involves converting nodes in a graph into lower-dimensional embeddings. Node embeddings are frequently utilized as features for downstream tasks. Therefore, the performance of these tasks is greatly influenced by the quality of the learned embeddings. Node representation learning methods are optimized using various loss functions, including (semi-)supervised, unsupervised, or self-supervised. (Semi-)supervised learning involves accessing ground truth information associated with a downstream task during training. However, when downstream class labels are available, the optimization may overfit to those labels and underutilize other more abundant graph signals, which can reduce generalization performance.

\begin{figure}[t]
\centering
\includegraphics[width=\linewidth]{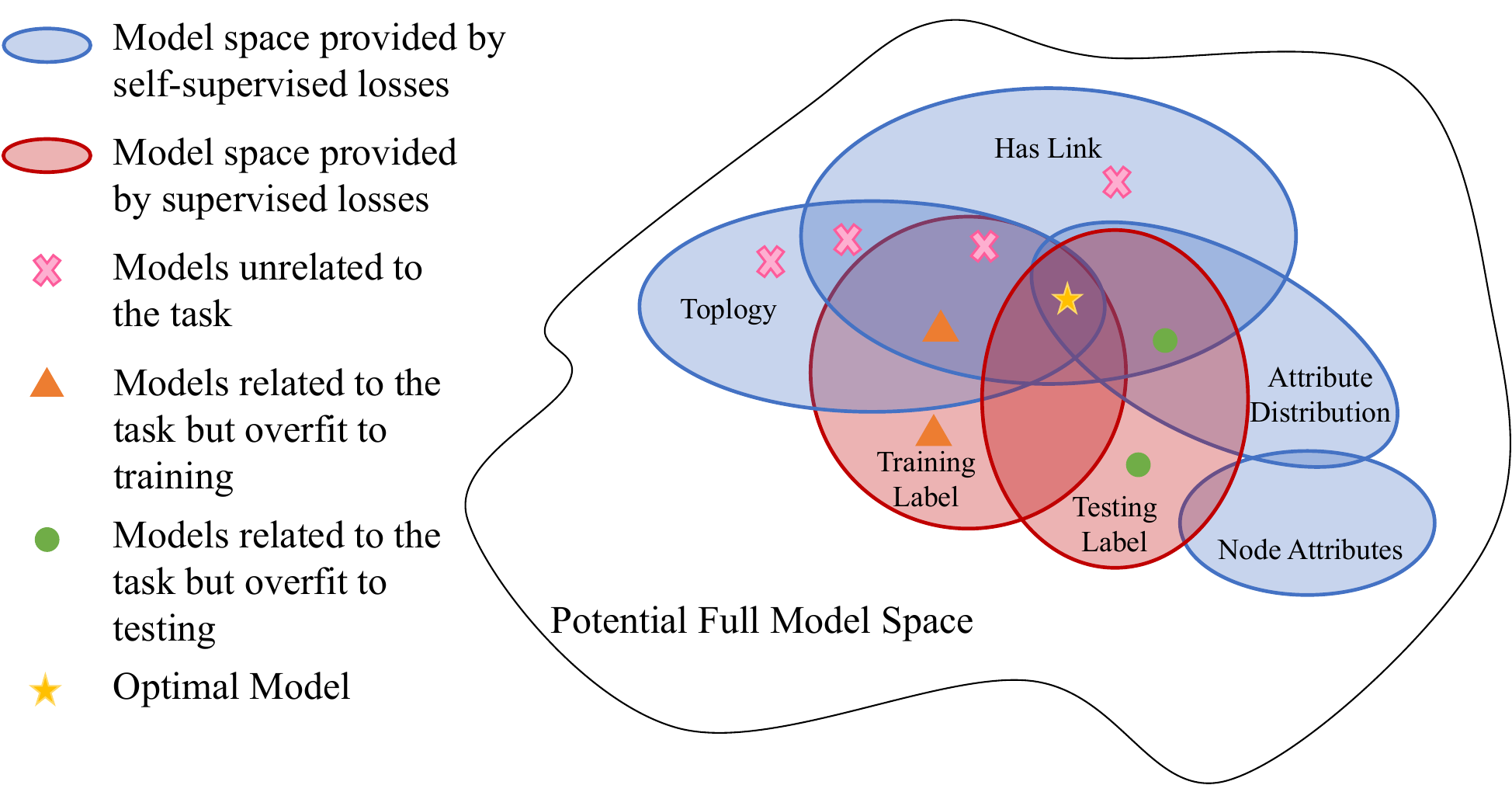}
\caption{Model space consisting of models trained by different graph signals (text labels in each model space). The optimal model for a downstream task is represented by a yellow star. Different graph signals may result in varying degrees of overfitting or generalization error. Our goal is to increase the likelihood of finding the optimal model by optimizing it through adapting the information learned from different graph signals.}
\label{fig:motivation}
\end{figure}

Unsupervised and self-supervised learning use the underlying patterns or graph structure for optimization. In this case, the node representations are learned to capture graph signals during training. However, the relationship between these graph signals and a downstream task is not always clear. Optimizing for irrelevant information can negatively impact task performance~\citep{bottle_neck}, as shown in Figure~\ref{fig:motivation_exp}.

While previous studies have demonstrated the importance of relevant self-supervised learning targets for enhancing downstream task performance~\citep{adgcl}, there has been limited research analyzing this relevance specifically for node representations. Although studies in graph pretraining have shown promise in leveraging task-relevant signals~\cite{pretrain1, mentorgnn}, they cannot be directly adapted to node-level tasks, creating a significant gap for node representations. Therefore, in this work, we propose a self-supervised approach, Task-Aware Contrastive Learning ({\sss}), which can actively adapt its contrastive loss function based on the downstream task. Given a pair/set of positive examples and negative examples, the general goal of contrastive loss is to bring positive examples together and negative examples apart from each other. Unlike self-supervised learning, {\sss} uses Task-Aware Contrastive Loss (\loss), which involves strategically selecting the positive examples that enhance the mutual information between the downstream task and node representations, such that {\loss} is relevant to the downstream task.

However, there are two main challenges in strategically selecting the positive examples that increase the mutual information between the downstream task and node representations. First, how each node label relates to a downstream task depends on its local neighborhood, e.g., a local subgraph, and on its node attributes. Instead of proposing a general positive example sampling strategy as in previous studies~\citep{sugrl,n2n}, one sampling strategy should be learned for {\em each node}. However, it's inefficient to learn one strategy for each node, especially when the size of the graph is large. Second, the amount of ground truth is usually limited. It's even more difficult to flearn a sampling strategy for each node if limited ground truth is available.

To tackle these challenges, we develop an XGBoost Sampler (\xgs) that uses limited ground truth labels in the training data to learn how to sample positive examples that increase the mutual information between the downstream task and node representations. Our use of {\xgs} not only increases the likelihood of sampling nodes to improve the relevance between the downstream task and node representations, but it also enhances the interpretability of models by indicating the importance of each graph signal for the downstream task. Our main contributions of this work are as follows:
\begin{itemize}
    \item We introduce a novel concept of supervision for representation learning (\sss), which integrates graph signals related to a downstream task into self-supervised optimization.
    \item We propose Task-Aware Contrastive Loss ({\loss}) as an example of {\sss} that aims to make the contrastive loss more relevant to a downstream task. It achieves this by utilizing {\xgs} to sample proper positive examples that enhance the mutual information between the downstream task and node representations. 
    \item We evaluate \loss\ with GNNs empirically and show that it can reduce generalization error and enhance performance significantly on both node classification and link prediction compared with state-of-the-art models.
\end{itemize}

\section{Notations and Preliminaries}

We first introduce basic concepts and define common notations in this section. A general graph is defined as $\gG=(\gV, \gE, \mX, \mY)$ where $\gV$ is the node set, $\gE:\gV\times\gV$ denotes all edges, $\mX\in\sR^{|\gV|\times m}$ represents node attributes ($m$ is the dimension), and $\mY=\{y_1, y_2, \cdots, y_{|\gV|}\}$ is the set of node labels in which the class label $y_u\in\{1, 2, \cdots, C\}$. The graph contains various {\em semantic relations} $\gR=\left\{r_1, r_2, \cdots, r_{|\gR|}\right\}$ which refer to relationships between pairs of nodes. Example relations include structural similarities such as shortest path distance or attribute similarities such as neighbor label distribution. The set of semantic relations used in our experiments is listed in Appendix~\ref{sec:semantic_relation}. We use these semantic relations as {\em graph signals} in this paper. For each relation $r\in\gR$, we define a similarity measure $s_r:\gV\times\gV\rightarrow [0,1]$ to record the strength of the relation between two nodes, and $S_r$ is the random variable describing the similarity values for $r$.

\subsection{Node Contrastive Learning}
Given a graph $\gG$, a graph-based model $f:\gG\rightarrow\sR^{|\gV|\times h}$ transforms nodes in the graph into node representations $Z=f(\gG)$, where $h$ is the embedding dimension. We denote $z_u$ as the embedding of node $u\in\gV.$ Specifically, $Z$ is optimized by Contrastive Loss (CL)~\citep{debias_cl, theory_cl, loco},  which is generally defined as:
\begin{small}
\begin{equation*}
    \mathcal{L}(u, v_p, \left\{v_{n_k}\right\}_{k=1}^{K})=
    -\log\frac{exp\left(z_u^{\top}z_{v_{p}}\right)}{exp\left(z_u^{\top}z_{v_{p}} \right)+\sum_{k=1}^{K}exp\left(z_u^{\top}z_{v_{n_k}} \right)},
\end{equation*}
\end{small}

\noindent where $u,v_{p},v_{n_k}\in\gV$. Here $u$ is a query node, $v_{p}$ is the positive example for $u$, and $v_{n_k}$ is a negative example. $K$ is the number of negative examples. The positive example $v_{p}$ is usually sampled from some probability function w.r.t $u$, e.g., based on the probability that an edge exists between $u$ and $v_p$~\cite{sugrl,gae} or the probability that $u$ and $v_p$ share a similar topological structure~\cite{n2n}. Conversely, negative examples are randomly sampled from $\gV$ or nodes distinct from $u$. If $B>1$ positive examples are sampled for each query node $u$, the contrastive loss becomes:
\begin{small}
\begin{align}\label{eq:cl}
    \mathcal{L}&\bigg(u, \left\{v_{p_b}\right\}_{b=1}^{B}, \left\{v_{n_k}\right\}_{k=1}^{K}\bigg)=\notag\\
    &-\frac{1}{B}\sum_{b=1}^{B}\log\Bigg[\frac{exp\left(z_u^{\top}z_{v_{p_b}}\right)}{exp\left(z_u^{\top}z_{v_{p_b}} \right)+\sum_{k=1}^{K}exp\left(z_u^{\top}z_{v_{n_k}}\right)}\Bigg]
\end{align}
\end{small}

\noindent The contrastive loss encourages positive examples to be embedded close to $u$ and discourages negative examples from being embedded close to $u$. 
Depending on the sampling strategy for positive examples, the loss can be categorized as {\em self-supervised}~\citep{n2n, gae, sugrl} or {\em supervised}~\citep{sup_cl} contrastive loss. 

\begin{definition} \textbf{Self-Supervised Contrastive Loss (SSCL).}\label{def:self_cl}
A graph-based model $f$ is typically optimized by SSCL with positive examples sampled from a selected subset of semantic relations $\gR_f\subseteq\gR.$ Specifically, for a particular relation $r$, $v_{p}^r$ is sampled from the set $\{v^{r}_{p} \:| \:s_{r}(u,v^{r}_{p})>\eta_{r}\}$ for $r$, where a threshold $\eta_{r}\in(0,1]$ is used to select examples with high similarity w.r.t $r$. 

When $B>|\gR_f|>1$, previous studies have typically sampled positive samples uniformly over the selected relations $\gR_f$, e.g.
\[
\left\{v_{p_b}\right\}_{b=1}^{B}=\cup_{r\in\gR_f}\{v_{p_i}^r\}_{i=1}^{\lfloor B/|\gR_f|\rfloor},
\] 
where each $v_{p_i}^r\in\{v^{r}_{p} \:| \:s_{r}(u,v^{r}_{p})>\eta_{r}\}.$ The negative examples are still sampled randomly from $\gV$ and then the sampled positive and negative examples are utilized to minimize the loss in \eqref{eq:cl}. The model space considered by various methods differs with respect to which relations the methods use for sampling positive nodes (i.e., $\gR_f$). This is illustrated by the blue regions in Figure~\ref{fig:motivation}. 
\end{definition}

\begin{definition} \textbf{Supervised Contrastive Loss (SCL).}\label{def:scl} In general, SCL uses ground truth labels to sample positive examples, 
e.g. 
\[
\left\{v_{p_b}\right\}_{b=1}^{B}\subset\{v \:| \:y_u=y_v, u,v\in V_L\},
\] 
where $V_{L}\subseteq\gV$ is the set of labeled nodes, and the negative examples are sampled randomly from $\gV$. Again the sampled positive and negative examples are utilized to minimize the loss in equation \eqref{eq:cl}. Then, the model space considered by SCL is depicted by the red (training) region in Figure~\ref{fig:motivation}. 
\end{definition}

Let $\gI(X_1;X_2)$ be the mutual information between two random variables $X_1$ and $X_2$. For SSCL, minimizing the loss in \eqref{eq:cl} with the equal number of positive examples, i.e., $\lfloor B/|\gR_f|\rfloor$, for each $r\in\gR_f$ is equivalent to uniformly maximizing $\gI(S_r;Z)~\forall r\in\gR_f$ \citep{infonce}.
For SCL, by sampling positive examples based on node labels $\mY$, minimizing the loss in \eqref{eq:cl} is equivalent to maximizing $\gI(\mY;Z)$~\citep{infonce}.

The learned node representations $Z$ can be used for a downstream task by learning another model that maps the node representation to a target class label $\mY$. For example, a classifier is a task-specific model that maps $Z$ to class labels for nodes.

\section{Learning Node Representations with Task-Aware Contrastive Loss}

In this section, we will provide motivation for Task-Aware Contrastive Learning ({\sss})  and then describe our implementation of Task-aware Contrastive Loss ({\loss}), which uses {\xgs} for {\sss}.

\subsection{Task-Aware Contrastive Loss}
\label{sec:task_cl}

As mentioned above, SCL is effectively maximizing $\gI(\mY;Z)$ and SSCL is effectively maximizing $\gI(S_r;Z)~\forall r\in\gR_f$. SCL will be more accurate for the downstream task $\mY$, but only if sufficient labeled data is available. SSCL is utilized when there is sparse label information available. However, we argue that the downstream performance of SSCL models will be impacted by whether uniformly maximizing $\gI(S_r;Z)~\forall r\in\gR_f$ is equivalent to maximizing $\gI(\mY;Z)$. Our observation is that there is a high chance that uniformly maximizing $\gI(S_r;Z)~\forall r\in\gR_f$ is {\em not} equivalent to maximizing $\gI(\mY;Z)$. Consequently, the learned node representations may be inadequate for the intended downstream task. We empirically verify this statement in Section~\ref{sec:label_perturb}.

To address this issue, we contend that the contribution of each $\gI(S_r;Z)$ for $r\in\gR_f$ to \eqref{eq:cl} should be weighted by the association between $r$ and the downstream task. To make the association between $\gR_f$ and the downstream task clearer, we define a function to assess it:
\begin{definition} 
\textbf{Task Positive Function}.\label{def:task_pos_func} The association between task $t$ and the set of semantic relations $\gR_f$ is quantified by the task positive function:
\begin{equation}\label{eq:task_pos_func}
    \sP_u\bigg(\sI_t(u,v)=1\bigg|\{s_r(u,v)|r\in\gR_f, v\in\gV\}\bigg),
\end{equation}
where $\sI_t(u,v)\in\{0,1\}$ is the indicator function to verify whether two nodes have the same feature w.r.t. to a particular task $t$. If $t$ is node classification, $\sI_t(u,v)=\sI\{y_u=y_v\}$. On the other hand, if $t$ is link prediction, $\sI_t(u,v)=\sI\{(u,v)\in\E\}$. If  $\sI_t(u,v)=1$, then $u$ and $v$ share the same task-relevant feature, or vice versa. The downstream task can be anything as long as $\sI_t(u,v)$ is defined to reflect the target of $t$.  
\end{definition}

\noindent Note that \eqref{eq:task_pos_func} is defined w.r.t. each node $u$. The strength of inter-node relationships is contingent upon the local subgraph of $u$, e.g., $\sP_u\big(\sI_t(u,v)=1\big\vert s_{r}(u,v)=c\big)\neq \sP_{u'}\big(\sI_t(u',v)=1\big\vert s_{r}(u',v)=c\big),$ where $u,u',v\in\gV$

By sampling positive examples based on \eqref{eq:task_pos_func}, we can ensure that the positive examples are relevant to task $t$ based on the distribution of $\{s_r(u,v)|r\in\gR_f\}$.  Therefore, our Task-Aware Contrastive Learning is defined as follows:
\begin{definition} \textbf{Task-Aware Contrastive Learning. (TCL)}\label{def:xgs}
\break TCL identifies positive examples of node $u$ in \eqref{eq:cl} based on, 
\begin{small}
\begin{equation}\label{eq:xgs}
    \left\{v_{p_b}\right\}_{b=1}^{B}=\argtopk_{\{v\}} \: \sP_u\bigg(\sI_t(u,v)=1\bigg|\big\{s_r(u,v)\big|r\in\gR_f, v\in\gV\big\}\bigg)
\end{equation}
\end{small}

\noindent where $k=B$ and $\argtopk$ returns the $\{v\}$ with top-$k$ probability values.
\end{definition}

We conjecture that sampling positive examples 
based on \eqref{eq:xgs} is similar to maximizing $\gI(\mY;Z)$, and therefore model generalization for downstream tasks will be improved. Based on the above hypothesis, our Task-aware Contrastive Loss ({\loss}) will actively adapt the sampling of positive examples for each node $u$, based on the pairwise similarity distributions for  $r\in\gR_f$.

To verify the hypothesis, we need to learn a function to model \eqref{eq:task_pos_func}. We propose {\em XGBoost Sampler} (\xgs), $f_u(v)$, to learn \eqref{eq:task_pos_func} by combining several weak classifiers learned from each relation $r\in\gR_f$. Then, $f_u(v)$ is used to approximate \eqref{eq:xgs} and sample positive examples such that minimizing \eqref{eq:cl} is similar to maximizing $\gI(\mY;Z)$.

Our proposed {\xgs} is an ensemble learning algorithm based on XGBoost~\citep{xgboost} that combines multiple simple regression stumps $f_{r,u}$ into one strong {\xgs} defined as: 
\begin{equation}\label{eq:f_xgs}
    f_u(v) = \sigma(\hat{y}_{u,v}) \;\;\; \mbox{ and } \;\hat{y}_{u,v}=\sum_{r=1}^{|\gR_f|}f_{r,u}(v).
\end{equation}
Here $\sigma:\sR\rightarrow[0,1]$ is a sigmoid function to convert arbitrary function values to probabilities, and $f_{r,u}$ is a relation regression stump for relation $r$ defined as:
\begin{definition}
\textbf{Relation Regression Stump ({\rtree}).} For semantic relation $r\in\gR_f$, a relation regression stump is
\begin{equation}\label{eq:rt}
    f_{r,u}(v)=
    \begin{cases}
            w_{r,0} & \text{ if } s_r(u,v)<\eta_{r,u}\\
            w_{r,1} & \text{ if } s_r(u,v)\geq\eta_{r,u}, 
    \end{cases}
\end{equation}

\noindent where $w_{r,0} \in \sR$, $w_{r,1} \in \sR$, and $\eta_{r,u}\in[0,1]$ is a similarity threshold.
\end{definition}
\noindent The learning target for each $f_u(v)$ is to model \eqref{eq:task_pos_func} for each $u\in\gV$.

However, with limited ground truth labels, it is  difficult to estimate \eqref{eq:task_pos_func} because $\sI_t(u,v)$ is unknown for most nodes. Also, learning \eqref{eq:task_pos_func} for all $u$ based on the pairwise similarity distributions for each $r\in\gR_f$ is computationally burdensome and time-consuming. To tackle these challenges, it is difficult to use the original learning algorithm from \cite{xgboost}. Therefore, the training algorithm of {\xgs} needs to be modified to adopt both local and global signals to learn $f_u(v)$ for each node.

We use the similarity threshold $\eta_{r,u}$ as a local signal to customize $f_{r,u}(v)$  for each $u$ and $r$. Specifically, the decision threshold $\eta_{r,u}$ is determined for each node depending on the local subgraph of $u$. In our experiments, we set $\eta_{r,u}$ to the $99$-percentile of the values of $\{s_r(u,v)\}$ given node $u$.

As mentioned earlier, some nodes in the graph may not have ground truth signals to learn $w_{r,0}$ and $w_{r,1}$ in $f_{r,u}(v)$, which means all $y_{u,v}$ is unknown for a node $u$. Additionally, even when a node has ground truth signals to learn, $f_{r,u}(v)$ is easily overfit to limited ground truth. Therefore, we tie the parameters $w_{r,0}$ and $w_{r,1}$ for all nodes and jointly learn them w.r.t. all available $y_{u,v}$ to improve model generalization. As such, $w_{r,0}$ and $w_{r,1}$ are learned from the global signals for each $r\in\gR_f.$ Specifically, $w_{r,0}$ and $w_{r,1}$ are the same for all nodes for {\rtree} related to $r$. The following section will show how to learn $\{w_{r,0}, w_{r,1}|r\in\gR_f\}$ for {\xgs}.

\subsection{XGboost Sampler Training}

\begin{algorithm}[t]
    \caption{XGBoost Sampler Training}\label{alg:xgboost}
    \begin{algorithmic}[1]
    \REQUIRE Ordered semantic relation set $\overline{\gR_f}=\{r_1, r_2, \cdots, r_{|\gR_f|}\}$, training instances $\gV_{L}\subset\gV$, $\{\eta_{r_{\tau},u}|r_{\tau}\in\overline{\gR_f}, u\in\gV\}$  
    \ENSURE $\{w_{r_\tau,0}, w_{r_\tau,1}|r_\tau\in\overline{\gR_f}\}$
    \STATE Initialize all $\hat{y}^{(0)}_{u,v}=0$
    \FORALL{$\tau\in \{1,2,\cdots, |\gR_f|\}$}
        \FORALL{$(u,v)\in \gV_{L}\times \gV_{L}$}
            \STATE $g_{u,v}\gets \sigma(\hat{y}^{(\tau-1)}_{u,v})-y_{u,v}$
            \STATE $h_{u,v}\gets \sigma(\hat{y}^{(\tau-1)}_{u,v})(1-\sigma(\hat{y}^{(\tau-1)}_{u,v}))$
        \ENDFOR
        \STATE $w_{r_\tau,0}\gets\eqref{eq:w0}$
        \STATE $w_{r_\tau,1}\gets\eqref{eq:w1}$
        \FORALL{$(u,v)\in \gV_{L}\times \gV_{L}$}
            \IF{$s_{r_\tau}(u,v)\geq\eta_{r_\tau, u}$} 
                \STATE $\hat{y}^{(\tau)}_{u,v}=\hat{y}^{(\tau-1)}_{u,v}+w_{r_\tau,1}$
            \ELSE
                \STATE $\hat{y}^{(\tau)}_{u,v}=\hat{y}^{(\tau-1)}_{u,v}+w_{r_\tau,0}$
            \ENDIF
        \ENDFOR
    \ENDFOR
    \end{algorithmic}
\end{algorithm}

Our XGBoost Sampler ({\xgs}) learns $\{w_{r,0}, w_{r,1}|r\in\gR_f\}$ iteratively, considering all nodes jointly. We first need to decide the training order of the $r\in\gR_f$ based on some metric, since searching over the best choice of $r$  for each $u$ during training as in \cite{xgboost} is challenging due to limited ground truth. Instead, we use a globally-specified training order for all nodes: $\overline{\gR_f}=\{r_1, r_2, \cdots, r_{|\gR_f|}\}$. We determine the order of relations based on $\sP_r(\sI_t(u,v) =1|f_{r,u}(v)=1)$ given the available ground truth labels. In our experiments, we order $\overline{\gR_f}$ in descending order (i.e., the relation with the highest task positive function value is listed as $r_1$). In Algorithm~\ref{alg:xgboost}, the training process starts at iteration $\tau=1$, and $w_{r_1,0}$ and $w_{r_1,1}$ are learned for $r_1\in\overline{\gR_f}.$ The process continues until $\tau=|\gR_f|$ and $\{w_{r_\tau,0}, w_{r_\tau,0}|r_\tau\in\overline{\gR_f}\}$ is output and used by all $f_u(v)$.

In the iteration $\tau$ (lines 3 to 15 in Algorithm~\ref{alg:xgboost}), $w_{r_{\tau},0}$ and $w_{r_{\tau},1}$ for $f_{r_\tau, u}$ are learned by minimizing:
\begin{small}
\begin{equation}\label{eq:xgs_loss}
    \mathcal{L}_{r_\tau} = \sum_{u\in \gV_{L}}\Bigg[\sum_{v\in \gV_{L}}\ell\left( y_{u,v},\;\hat{y}^{(\tau-1)}_{u,v} + f_{r_{\tau},u}(v)\right)+\Omega(f_{r_{\tau},u})\Bigg],
\end{equation}
\end{small}

\noindent where $ \hat{y}^{(\tau-1)}_{u,v}= \sum_{i=1}^{\tau-1}f_{r_i,u}(v)$, $y_{u,v}=\sI_t(u,v)$, $\Omega(f_{r_{\tau},u})=\lambda(w^2_{r_{\tau},0}+w^2_{r_{\tau},1})/2$, $\lambda\in\sR$, and $\gV_{L}$ is determined based on the training set, i.e., labeled nodes in classification or nodes having edges in link prediction. $\ell(\cdot, \cdot)$ here is the binary cross-entropy function. By optimizing \eqref{eq:xgs_loss}, we greedily add $f_{r_{\tau},u}$ to improve our final loss.

A second-order approximation can be used to optimize \eqref{eq:xgs_loss} and determine $w_{r_{\tau},0}$ and $w_{r_{\tau},1}$ for $f_{r_{\tau},u}$:

\begin{small}
\begin{align*}
    \mathcal{L}_{r_\tau} \simeq & \sum_{u\in U_{L}}\sum_{v\in U_{L}}\Big[\ell\left(y_{u,v}, \hat{y}^{(\tau-1)}_{u,v}\right) \\
    &+ g_{u,v}f_{r_{\tau},u}(v)+\frac{1}{2}h_{u,v}f^{2}_{r_{\tau},u}(v)\Big]+\sum_{u\in U_{L}}\Omega(f_{r_{\tau},u}),\\
    g_{u,v}&=\partial_{\hat{y}^{(\tau-1)}_{u,v}}\ell(y_{u,v}, \hat{y}^{(\tau-1)}_{u,v})=\sigma(\hat{y}^{(\tau-1)}_{u,v})-y_{u,v},\\
    h_{u,v}&=\partial^2_{\hat{y}^{(\tau-1)}_{u,v}}\ell(y_{u,v}, \hat{y}^{(\tau-1)}_{u,v})=\sigma(\hat{y}^{(\tau-1)}_{u,v})(1-\sigma(\hat{y}^{(\tau-1)}_{u,v})).
\end{align*}
\end{small}
$\ell\left(y_{u,v}, \hat{y}^{(\tau-1)}_{u,v}\right)$ can be dropped because it does not affected by $w_{r_{\tau},0}$ and $w_{r_{\tau},1}$. $\mathcal{L}_{r_\tau}$ becomes
\begin{small}
\begin{align*}
    \widetilde{\mathcal{L}}_{r_\tau} =& \sum_{u\in U_{L}}\sum_{v\in U_{L}}\Bigg[g_{u,v}f_{r_{\tau},u}(v)+\frac{1}{2}h_{u,v}f^{2}_{r_{\tau},u}(v)\Bigg]+\sum_{u\in U_L}\Omega(f_{r_{\tau},u})
\end{align*}
\end{small}
Let $I_{u,0}=\{v|s_{r_{\tau}}(u,v)<\eta_{r_{\tau}, u},v\in\gV_L\}$ and $I_{u,1}=\{v|s_{r_{\tau}}(u,v)\geq\eta_{r_{\tau}, u},v\in\gV_L\}$. The above equation can be re-written as

\begin{small}
\begin{align*}
    \widetilde{\mathcal{L}}_{r_\tau} = \sum_{u}\Bigg[&\sum_{v\in I_0}g_{u,v}w_{r_{\tau},0}+\frac{1}{2}h_{u,v}w^2_{r_{\tau},0}+\sum_{v\in I_1}g_{u,v}w_{r_{\tau},1}\\
    &+\frac{1}{2}h_{u,v}w^2_{r_{\tau},1}+\frac{1}{2}\lambda(w^2_{r_{\tau},0}+w^2_{r_{\tau},1})\Bigg]
\end{align*}
\end{small}

The $w_{r_{\tau},0}$ and $w_{r_{\tau},1}$ that minimize $\widetilde{\mathcal{L}}_{r_\tau}$ are derived by solving $\partial_{w_{r_{\tau},0}}\widetilde{\mathcal{L}}_{r_\tau}=0$ and $\partial_{w_{r_{\tau},1}}\widetilde{\mathcal{L}}_{r_\tau}=0$:

\begin{equation}
    w_{r_{\tau},0}=\frac{-\sum_{u}\sum_{v\in I_{u,0}}g_{u,v}}{\sum_{u}(\sum_{v\in I_{u,0}}h_u,v)+\lambda}\label{eq:w0}
\end{equation}
\begin{equation}
    w_{r_{\tau},1}=\frac{-\sum_{u}\sum_{v\in I_{u,1}}g_{u,v}}{\sum_{u}(\sum_{v\in I_{u,1}}h_u,v)+\lambda}\label{eq:w1},
\end{equation}

Thus, $w_{r_{\tau},0}$ and $w_{r_{\tau},1}$ of $f_{r_{\tau},u}(v)$ can be calculated from \eqref{eq:w0} and \eqref{eq:w1}. As we will show later in Section~\ref{sec:computation}, {\xgs} is an efficient approach in practice compared to learning XGBoost based on \cite{xgboost}.

By combining all $\{w_{r_\tau,0}, w_{r_\tau,1}|r_\tau\in\overline{\gR_f}\}$ and $\{\eta_{r_{\tau},u}|r_\tau\in\overline{\gR_f}\}$ for $u$, {\xgs} can classify whether a node $v$ is the positive example for $u$ based on $f_u(v)$ in \eqref{eq:f_xgs}. Moreover, $w_{r_\tau,0}$ and $w_{r_\tau,1}$ associated with each $f_{r_\tau,u}$ offer valuable insights into the significance of relation $r_\tau$ with respect to the downstream task. Higher values of $w_{r_\tau,0}$ and $w_{r_\tau,1}$ indicate a greater influence in the determination of whether a node qualifies as a positive example. By deriving these weights for all semantic relations, {\xgs} enhances the interpretability of node representations.

\subsection{Computation Complexity Analysis}
\label{sec:time}
The process for learning node representations with {\loss} includes the time for (i) calculatint similarities, (ii) learning {\xgs}s, (iii) sampling positive examples for all nodes, and (iv) the time for training a GCN model with the sampled positive examples.

For (i), we calculate similarity matrices once and store them in a database (i.e., there is no need to recalculate on each iteration of training node representations), and the complexity depends on the specific relation. Some semantic relations (e.g., Link) are $O(|\gE|)$ if they can be calculated and stored in sparse matrices. Other semantic relations (e.g., Attr Sim.) require $O(|\gV|^2)$ because they consider full  pairwise node information. To make these relations more efficient, some methods only focus on node pairs with edges and thus use a sparse matrix implementation to reduce  complexity to $O(|\gE|)$\cite{gcn}.

For (ii-iii), a naive implementation will be   $O(|\gV|^2)$ because, for arbitrary graph relations, Algorithm 1 will need to consider all node pairs. Note that $|\gR_f|$ is a constant, which we assume does not grow with graph size. However, by tying the $w_{r,0}$ and $w_{r,1}$ parameters, both training and inference for {\xgs} can be implemented efficiently with a GPU-based approach to greatly reduce computation time~\cite{cuda}. According to empirical computational results in Section~\ref{sec:computation}, our {\xgs} is much more efficient than $O(|\gV|^2)$ in practice. Additionally, for (iii), we need to sample positive examples from all nodes for each query node $u$, e.g., checking $s_r(u,v)>\eta_{r,u}$ for all $v\in\gV$, which is $O(|\gV|^2)$. We can reduce the complexity to $O(|\gE|)$ if we only sample positive examples from the nodes with edges gather and scatter operations. Because the computational complexity is proportional to the number of pairs with gather and scatter operations~\cite{pyg}, and the number of pairs is no greater than the number of edges.

The complexity of (iv) is $O(|\gV|)$ when the node representation is learned with a fixed number of positive nodes $B$ and negative nodes $K$ to optimize Eq. 1. With gather and scatter operations, the computation time is proportional to the number of pairs ($|\gV|\times(B+K)$). Because $B+K$ is constant, the complexity of (iv) is $O(|\gV|)$. 

Overall, this means TCL is worst case $O(|\gV|^2)$, but there are several implementation approximations that can be used to make the full process subquadratic.

\section{Performance Evaluation}
\begin{table*}[t]
\caption{Accuracy of Node Classification Task. Statistically significant results are marked with * (p-value$<0.05$). The top scores are marked with \first{bold blue}, the 2nd scores are marked with \second{blue}, the 3rd scores are marked with \third{violet}.}
\vspace{-4mm}
\setlength{\tabcolsep}{2.3pt}
\begin{center}
\begin{tabular}{lllllll}
\hline
Method & Cora & CiteSeer & PubMed & Computers & Photo \\
\hline
DGI &84.42±0.17 & 69.37±2.1 & 84.35±0.08 & 83.59±0.2 & 91.92±0.06 \\
SUGRL & 82.86±0.42 & 72.59±0.19 & 85.17±0.16 & 85.28±1.11 & 92.85±0.09 \\
ASP & \second{84.58±0.35} & \second{73.53±0.01} & 84.0±0.01 & 84.03±0.01 & 90.52±0.01 \\
PolyGCL & 65.62±0.07 & 71.10±0.01 & 85.28±0.01 & 82.32±0.01 & 89.92±0.01 \\
\hline
GCN & 82.01±0.39 & 68.73±0.29 & \first{86.65±0.07}* & 88.01±0.18 & 92.59±0.1 \\
GAT & 83.01±0.36 & 69.16±0.31 & 85.08±0.11 & \third{88.16±0.12} & \third{93.0±0.08} \\
DGCN & 80.21±0.27 & 69.96±0.21 & 83.57±0.1 & 86.23±0.38 & 90.87±0.17 \\
MetaPN & 83.96±0.39 & 73.02±0.16 & 85.32±0.13 & 78.24±0.29 & 88.72±0.63 \\
CoCoS & 83.1±0.87 & 72.65±0.18 & \second{86.30±0.22} & 83.99±0.65 & 92.07±0.16 \\
\hline
XTCL(GCN) & \first{85.19±0.13}* & \first{73.55±0.14} & \third{86.09±0.09} & \first{89.54±0.09}* & \first{93.84±0.09}* \\
XTCL(GAT) &  \third{84.13±0.24} & \third{73.26±0.32} & 85.47±0.08 & \second{88.50±0.10} & \second{93.15±0.12} \\
\hline
\end{tabular}
\label{tab:nc}
\end{center}
\end{table*}

Our proposed model is Task-Aware CL with {\xgs}, which we refer to as XTCL. Note that we can optimize any GNN method with {\loss}. In this work, we demonstrate the effectiveness of our {\loss} with GCN~\citep{gcn} and GAT~\citep{gat}. The dataset descriptions are in Table~\ref{tab:dataset}. In our empirical evaluation, we aim to answer:
\begin{itemize}
    \item {\bf RQ1.} Does {\loss} improve performance on node classification and link prediction compared to other SOTA methods?
    \item {\bf RQ2.} How do supervision signals and training data size impact GNN performance? 
    \item {\bf RQ3.} Is it important to make the loss function task-aware? 
    \item {\bf RQ4.} What semantic relations are important for each downstream task?
\end{itemize}

\noindent\textbf{Dataset and Baseline} We use widely used datasets to report the results of node classification and link prediction. Cora, CiteSeer, and PubMed are citation networks~\citep{cora}, and Photo and Computers are product networks~\citep{amazon}. If two products are often bought together, there is a link between them. The details of the dataset statistics are shown in Table~\ref{tab:dataset}. For baseline methods, we compare our {\sss} with different training signals. DGI~\citep{dgi}, SUGRL~\citep{sugrl}, ASP~\citep{asp}, and PolyGCL~\citep{polygcl} are trained with their specified unsupervised or self-supervised signals. Task-specific models are trained for these methods to predict targets for downstream tasks. Variations of GNN models: GCN~\citep{gcn}, GAT~\citep{gat}, DGCN~\citep{dgcn}, MetaPN~\citep{metapn}, CoCoS~\citep{cocos}, and SubGraph~\citep{subgraph} are trained with supervised signals based on their downstream tasks. See Appendix~\ref{apx:exp_env} to learn more about the detailed settings.

\begin{small}
\begin{table}[h]
\caption{Data Statistics. Cora, CiteSeer and PubMed are from \cite{cora}. Computers and Photo are from \cite{amazon}.}
\vspace{-4mm}
\setlength{\tabcolsep}{2pt}
\label{tab:dataset}
\begin{center}
\begin{tabular}{lrrrr}
\hline
Dataset & Node & Edge & Node Class & Node Attr. \\
\hline
Cora & 2708 & 10556 & 7 & 1433 \\
CiteSeer & 3327 & 9104 & 6 & 3703 \\
PubMed & 19717 & 88648 & 3 & 500 \\
Photo & 7650 & 238162 & 8 & 745 \\
Computers & 13752 & 491722 & 10 & 767 \\
\hline
\end{tabular}
\end{center}
\end{table}
\end{small}

\subsection{RQ1. Performance on Downstream Tasks}
\label{sec:task_results}

\noindent\textbf{Node Classification}\\
The performance results are shown in Table~\ref{tab:nc}. XTCL(GCN) and XTCL(GAT) are GNNs trained with the proposed {\loss}. Our {\xgs} uses the available training labels to learn a model for each $u\in\gV$, which we use to select positive examples for {\sss}. For node representations that are generated in {\loss} (self-supervised or unsupervised settings), we use Logistic Regression models as the final classifiers for label assignment (with learned embeddings as input). We use $10\%$ of the labels for training and $90\%$ of the labels for testing, and the reported results are the average of five runs.

Table~\ref{tab:nc} shows that, with only a small number of training labels ($10\%$) to train {\xgs}, XTCL(GCN) significantly outperforms most of the state-of-the-art models in both unsupervised and \break (semi-)supervised settings. XTCL(GCN) achieves slightly lower performance compared to GCN in PubMed, and this may result from the smaller number of class label values (3 in Table~\ref{tab:dataset}), which makes GCN generalize to test data easily. (See Appendix~\ref{apx:gcn_error} for a more detailed explanation w.r.t. PubMed.) The overall results in the table demonstrate that minimizing {\loss} is similar to maximizing $\gI(\mY;Z)$ and significantly improves the performance on node classification.

\begin{table*}[t]
\caption{AUC of Link Prediction. Statistically significant results are marked with * (p-value$<0.05$). The top scores are marked with \first{bold blue}, the 2nd scores are marked with \second{blue}, the 3rd scores are marked with \third{violet}.}
\begin{center}
\begin{tabular}{lllllll}
\hline
Method & Cora & CiteSeer & PubMed & Computers & Photo \\
\hline
DGI & \third{93.36±0.2} & 94.17±0.21 & 94.18±0.09 & 84.0±4.58 & 93.03±0.08 \\
SUGRL  & 93.74±0.16 & \third{96.89±0.09} & \third{95.81±0.06} & 91.89±0.26 & 95.50±0.06 \\
ASP  & 91.45±0.01 & 95.30±0.01 & 93.47±0.01 & 89.02±0.01 & 91.19±0.02 \\
PolyGCL  & 70.48±0.01 & 64.65±0.01 & 76.40±0.04 & 86.00±0.01 & 85.13±0.01 \\
\hline
GCN  & 92.9±0.16 & 95.97±0.12 & 95.33±0.07 & 95.59±0.06 & 96.9±0.02 \\
GAT  & 93.0±0.17 & 95.28±0.21 & 95.38±0.06 & 95.05±0.07 & 96.82±0.03 \\
DGCN   &  91.56±0.16 & 89.46±4.45 & 95.36±0.11 & 93.73±0.05 & 95.64±0.42 \\
SubGraph & 87.01±1.33 & 88.95±2.20 & 95.08±0.15 & \first{98.00±0.05}* & \second{98.21±0.02} \\
\hline
XTCL(GCN) & \first{94.07±0.13}* & \first{97.13±0.06}* & \first{96.72±0.06}* & \second{97.32±0.03} & \first{98.38±0.08}* \\
XTCL(GAT)& \second{93.95±0.13} & \second{97.01±0.07} & \second{96.64±0.03} & \third{96.43±0.06} & \third{97.44±0.05} \\
\hline
\end{tabular}
\label{tab:link}
\end{center}
\end{table*}

\noindent\textbf{Link Prediction}\\
For link prediction tasks, we compare to similar methods. We remove MetaPN and CoCoS because they are semi-supervised models for node classification and cannot be adapted for link prediction. We follow the random split strategy used by \citep{subgraph}. To ensure the training data is limited, we use $60\%$ of the links for training and $40\%$ of the links for testing and report the experiment results, averaged over five runs, in Table~\ref{tab:link}. The observations from Table~\ref{tab:link} are similar to those of Table~\ref{tab:nc}. By training GCN with XTCL, its performance can be improved significantly also for link prediction. This indicates that, by increasing the probability of selecting proper positive examples for link prediction, {\sss} successfully improves performance over other methods.

\subsection{RQ2. Comparing Various Supervision Signals}
\label{sec:comp_signals}

In this experiment, we explore whether {\xgs} can improve \loss\ performance on node classification as the training label size increases. We add two new methods for comparison: Ceiling Supervised Contrastive Loss (Ceiling SCL) and Naive Task-Aware CL (Naive TCL). See definitions below.

\begin{definition}
{\bf Ceiling Supervised Contrastive Loss (Ceiling SCL).}\label{def:cscl}
We use Ceiling SCL to refer to a method that uses ground truth labels to identify positive examples, i.e. 
\[\left\{v_{p_b}\right\}_{b=1}^{B}\subseteq\{v \:| \:\sI_t(u,v)=1, v\in V\} \]
The difference between SCL in Def.~\ref{def:scl} and Ceiling SCL is that Ceiling SCL can utilize the label of $y_u$ even when $u$ is not in $V_L$.
\end{definition}

\begin{definition} \textbf{Naive Task-Aware CL (Naive TCL).}\label{def:ntcl}
Naive TCL is a combination of SSCL in Def.~\ref{def:self_cl} and SCL in Def.~\ref{def:scl}. It uses Def.~\ref{def:self_cl} to select positive examples when $u\notin V_L$
If $u,v \in V_L$, the positive examples are sampled from the set $\{v \:|\:\sI_t(u,v)=1, u,v\in V_{L}\}$.
\end{definition}

\begin{figure}
    \centering
    \includegraphics[width=0.7\linewidth]{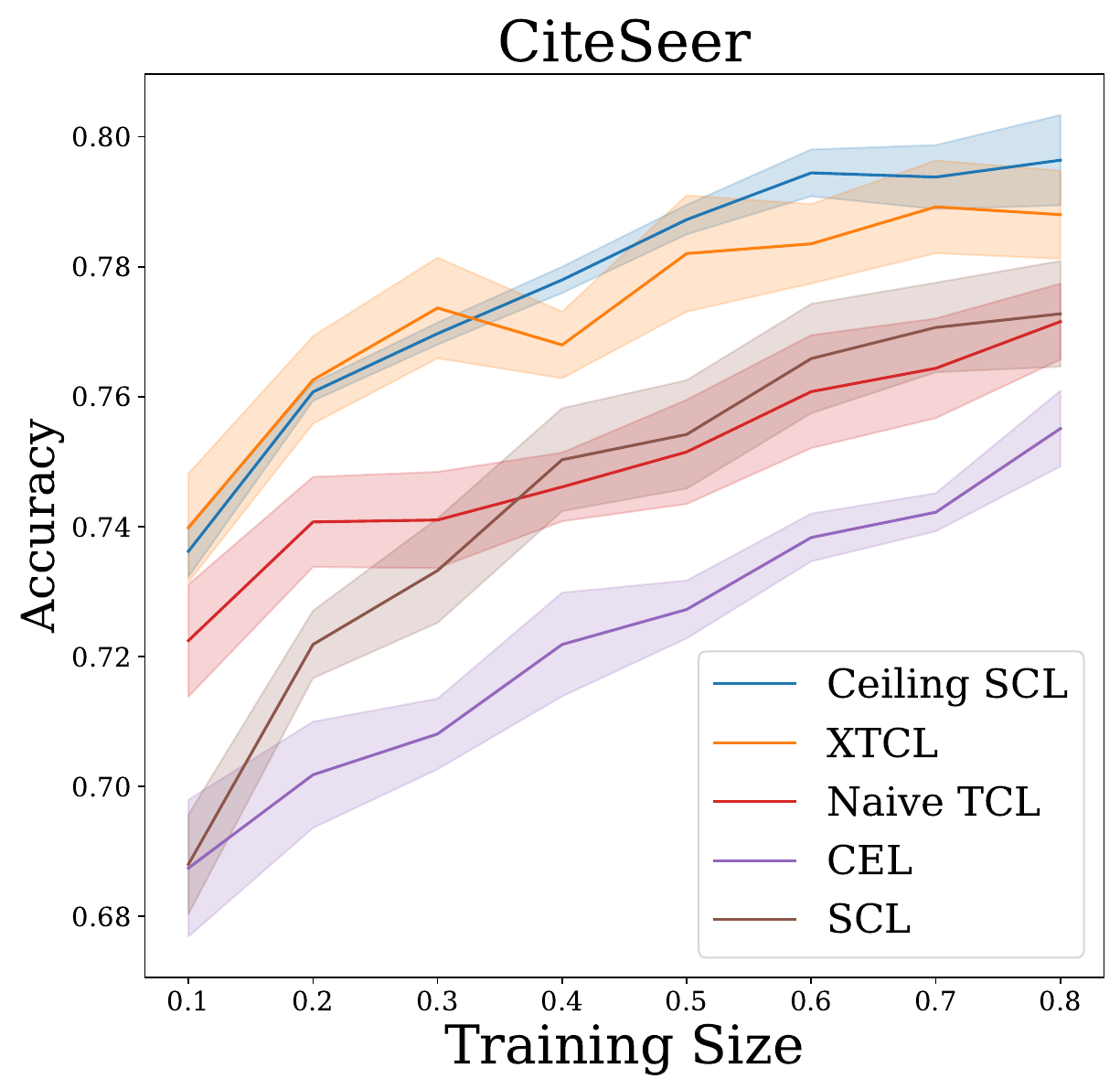}
    \caption{Learning curves of GCNs trained with various supervision signals. Our {\xgs} is effective because {\loss} continues to outperform others when the training size increases.}
    \label{fig:sup_vs_our}
\end{figure}

The results are shown in Figure~\ref{fig:sup_vs_our}. We use the same $20\%$ testing labels for all evaluated models, and the results are the average of five runs. We train GCN with various objective functions, and Logistic Regression is used as the classifier for all CL models. 

Figure~\ref{fig:sup_vs_our} shows that XTCL outperforms Naive TCL because {\xgs} increases the probability that $\sI_t(u,v)=1$ for a positive example $v$. However, as the training size increases, the performance gap between XTCL and Naive TCL becomes smaller. This is because Naive TCL has access to more training labels, allowing it to sample more positive nodes for which $\sI_t(u,v)=1$ as positive examples. A similar observation can be applied to SCL. Furthermore, XTCL can sometimes be comparable to the Ceiling SCL because {\loss} enhances model generalization further by incorporating abandoned graph signals during node representation learning.

\subsection{RQ3. Importance of Task-Aware Loss Function}
\label{sec:label_perturb}

In Section~\ref{sec:task_cl}, we briefly mentioned that the selection of the loss function will impact the model generalization for a downstream task. Our main argument is that minimizing the value of a loss function is {\em not} equivalent to maximizing $\gI(\mY;Z)$. Consequently, model performance and generalization degrade severely. 

\begin{figure}[h]
    \centering
    \begin{subfigure}{.5\linewidth}
      \centering
      \includegraphics[trim={0 0 0 0},clip,width=\linewidth]{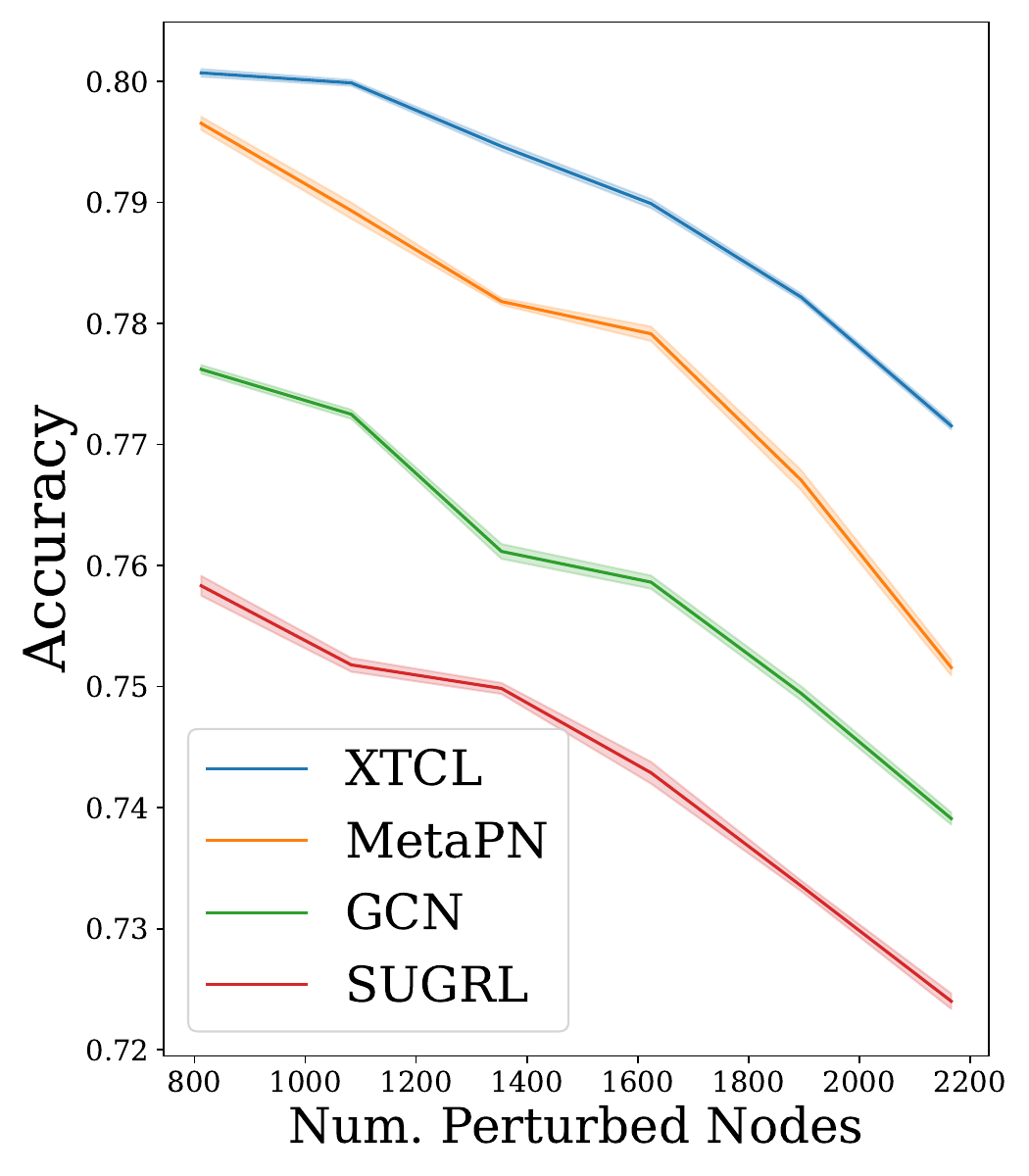}
      \caption{Node Classification ACC.}
      \label{fig:perturb_attr_dist}
    \end{subfigure}%
    \begin{subfigure}{.5\linewidth}
      \centering
      \includegraphics[trim={0 0 0 0},clip,width=\linewidth]{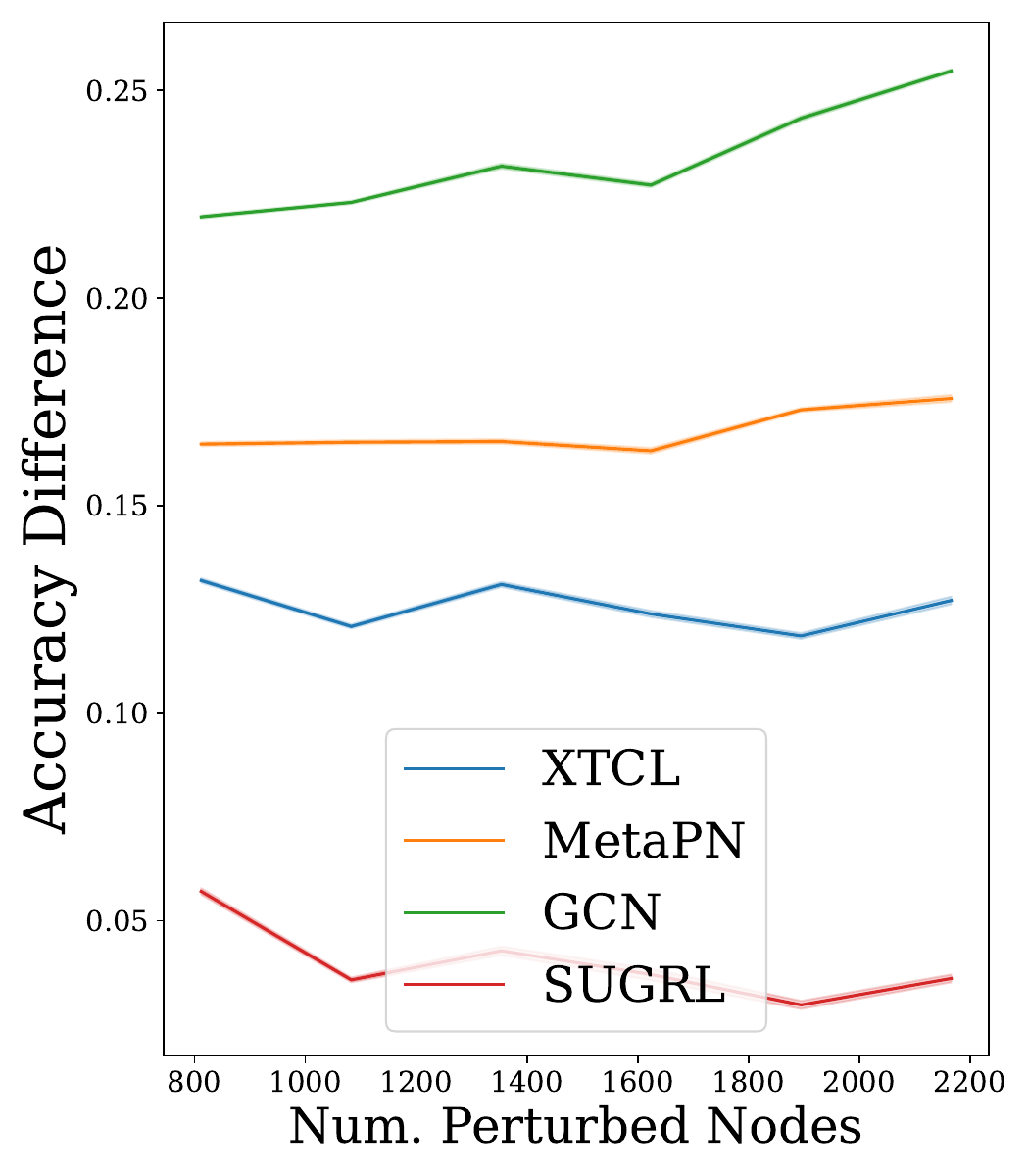}
      \caption{Train ACC - Test ACC.}
      \label{fig:acc_diff_attr_dist}
    \end{subfigure}
\caption{Results on semi-synthetic data when varying the dependency between node labels and a graph signal (1-Hop Attr. Dist.). The results show that minimizing losses, which are not task-aware, is {\em not} equivalent to maximizing the mutual information $\gI(\mY;Z)$, which decreases model generalization on node classification.}
\label{fig:motivation_exp}
\end{figure}

To empirically investigate the above conjecture, we conducted an experiment using semi-synthetic data generated from Cora. The downstream task is a classification task. First, we choose a semantic relation $\hat{r}$ subject to the condition that knowing $s_{\hat{r}}(u,v)$ does not inform the probability of $y_u=y_v$, and $\gI(S_{\hat{r}};Z)$ is less affected by optimizing the loss function that generates $Z$. Then, we gradually perturb some node labels in $\mY$ such that $\gI(\mY';S_{\hat{r}})>\gI(\mY;S_{\hat{r}})$. $\mY'$ is the new node label set with some labels perturbed. This will create a situation in which minimizing the value of a loss function is not equivalent to maximizing $\gI(\mY';Z)$. Because $\gI(S_{\hat{r}};Z)$ remains similar by optimizing the loss function, $\gI(\mY';Z)$ drops when $\gI(\mY';S_{\hat{r}})$ increases. We want to observe how this situation affects  model performance. When the number of perturbed nodes increases, $\gI(\mY';S_{\hat{r}})$ will increases accordingly in Figure~\ref{fig:motivation_exp}. The $\hat{r}$ for this experiment is 1-Hop Attr Dist. The details of label perturbation process and the experiment settings of Figure~\ref{fig:motivation_exp} are in Appendix~\ref{apx:label_perturb}.

We compare GNNs optimized by these loss functions: self-supervised (SUGRL~\citep{sugrl}), supervised (GCN~\citep{gcn}), semi-supervised (MetaPN~\citep{metapn}), and our task-aware contrastive loss (XTCL). The goal is to observe how the performance of different models is affected when $\gI(\mY';S_{\hat{r}})$ (number of perturbed nodes) increases.

\begin{figure*}[t]
  \centering
  \includegraphics[width=\linewidth]{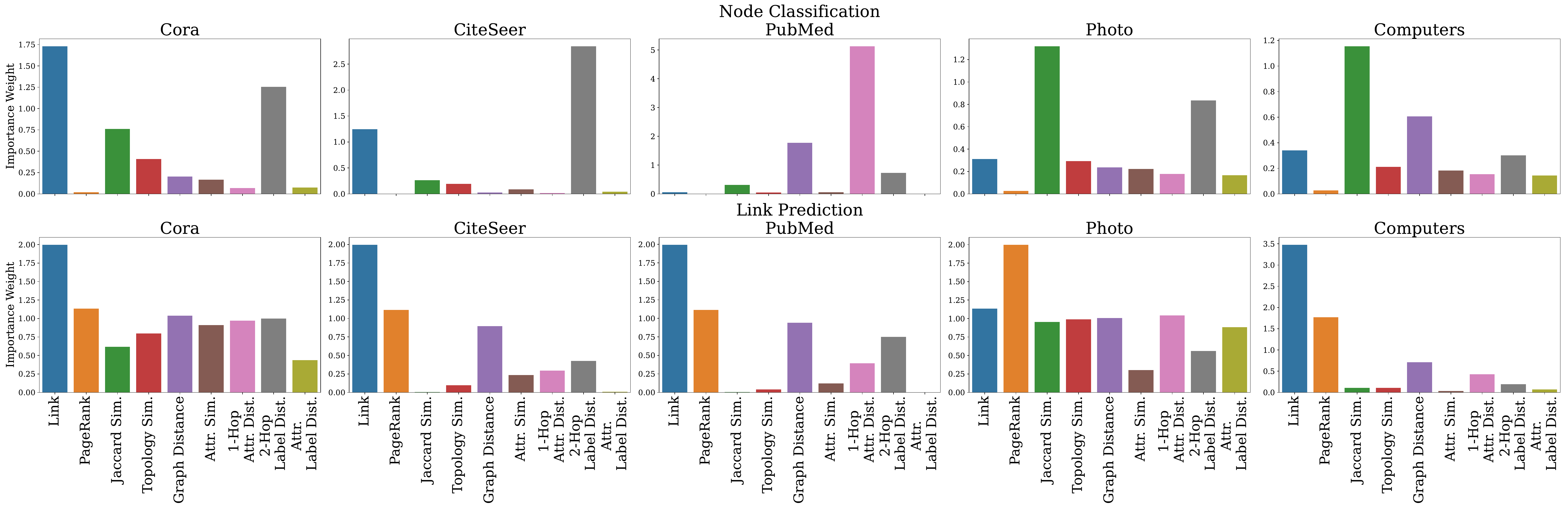}
  \caption{Importance Weights of Semantic Relations. This figure demonstrates the importance of these graph signals (x-axis) to each downstream task and how our {\xgs} adapt {\loss} to improve model performance.}
  \label{fig:weight}
\end{figure*}

Both self-supervised methods (SUGRL) and semi-supervised methods (MetaPN) have greater performance degradation because the generated representation does not capture 1-Hop Attr Dist by maximizing their loss functions in the embedding space. This illustrates that, since maximizing $\gI(S_r;Z)$ does not maximize $\gI(\mY';Z)$ simultaneously, the performance drops severely. As supervised methods (GCN), their performance is subject to the data distributions in training and testing, which is the overlapping area between the two red circles in Figure~\ref{fig:motivation}. GCN in Figure~\ref{fig:motivation_exp} overfits to training data because the number of training labels is insufficient (10\% of the data). The severity of GCN's overfitting is shown in Figure~\ref{fig:acc_diff_attr_dist}.

In contrast, our {\loss} shows relatively robust performance in Figure~\ref{fig:motivation_exp}. Because XTCL can actively adapt the number of positive examples for each $r\in\gR_f$ w.r.t. the downstream task, minimizing {\loss} is now similar to maximizing $\gI(\mY;Z)$. Therefore, the generated node representations can generalize better w.r.t. downstream task and largely reduce the chance of overfitting to the training data.

\subsection{Importance Weights of Semantic Relations}
The weights $w_{r,0}$ and $w_{r,1}$ learned from the \xgs\ indicate the importance of semantic relation $r$ in determining whether a node is a positive example for $u$. This weights also show the importance of each $r$ to a downstream task. The importance weight in Figure~\ref{fig:weight} is $\max(|w_{r,0}|, |w_{r,1}|)$ for each relation. The definition of each semantic relation is in Section~\ref{sec:semantic_relation}. Figure~\ref{fig:weight} demonstrates that different tasks (e.g., node classification vs. link prediction) rely on different semantic relations to determine positive examples, which means one semantic relation that is important to a downstream task may not be as critical as another downstream task. This result corresponds to our main argument that uniformly maximizing $\gI(S_r;Z)~\forall r\in\gR_f$ is not always equivalent to maximizing $\gI(\mY;Z)$. Figure~\ref{fig:weight} also demonstrates that {\xgs} improves the interpretability of the learned embeddings by illustrating which semantic relation is critical for a downstream task.

\begin{figure}
    \centering
    \includegraphics[width=0.7\linewidth]{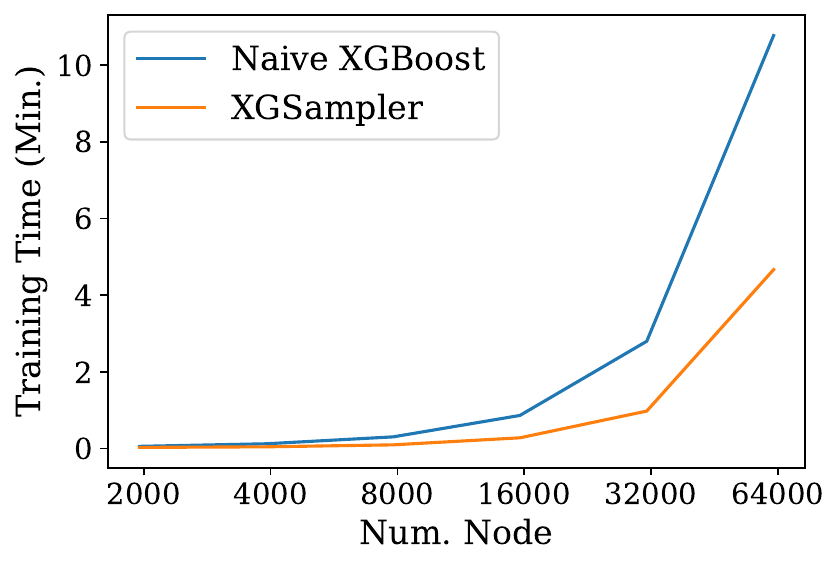}
    \caption{Training Time for Random Graph as graph size increases.}
    \label{fig:time}
\end{figure}

\subsection{Empirical Training Time Analysis}
\label{sec:computation}

In this experiment, we empirically demonstrate the $O(|\gV|^2)$ of our training process (when we consider full pairwise graph relations without approximation) compared to a naive implementation of XGBoost.
We use a random graph generated according to \cite{gcn}. The number of edges is set to $2|\gV|$. The labels $\gY$ and $\gX$ are randomly assigned. In our implementation, we first learn {\xgs}s for all nodes, e.g., $\{f_u(v)|u\in\gV\}$. Then, we use $f_u(v)$ to sample positive examples for for each $u\in\gV$. Finally, a GCN model is trained based on the sampled positive examples with \eqref{eq:cl}. According to the results in Figure~\ref{fig:time}, the average computational time for learning {\xgs}s is reduced by $60\%$ on average with GPU implementation, compared to Naive XGBoost. 


\section{Related Work}

Contrastive Learning has been applied to learning representations for various types of data. The selection of appropriate positive and negative examples is a crucial step in optimizing contrastive loss. In the context of Computer Vision (CV), various image augmentation techniques such as rotation and clipping have been proposed to generate positive examples from the query image~\citep{cl_review, simclr}. Conversely, in the field of Neural Language Processing (NLP), several pre-trained models are employed to generate positive samples for words or sentences~\citep{simcse}. These data augmentation approaches have demonstrated significant improvements in CV and NLP tasks.

Numerous task-aware contrastive objectives have been proposed for word and image representations. The initial work is Supervised Contrastive Learning ~\citep{sup_cl}, which selects positive and negative examples based on the available labels in the training data. The performance of text ~\citep{word_task_cl,negative_weights} and image classification tasks~\citep{sup_cl} has been enhanced through the use of supervised contrastive loss. The SCL depicted in Figure~\ref{fig:motivation_exp} and Figure~\ref{fig:sup_vs_our} adopts a similar setting to supervised contrastive learning, and both figures illustrate that the performance is constrained by unlabeled nodes due to the absence of an appropriate method for generating positive samples for those nodes.

Several graph augmentation techniques, e.g., adding/dropping edges~\citep{adgcl, infograph, pretrain1}, or modifying attributes~\citep{dgi, graph_cl}, have been proposed.  However, these graph augmentation techniques are specifically designed for entire graphs, whereas a single node is more sensitive to local changes made to the graph through these techniques. Consequently, many methods are proposed to sample positive examples based on structural or topological similarity in the graphs~\citep{sugrl,n2n}. However, the relationship between their unsupervised contrastive signals and the downstream task remains unclear, which may lead to arbitrary performance deterioration, as demonstrated in  Table~\ref{tab:nc} and Table~\ref{tab:link}.


\section{Conclusion} 
To ensure that minimizing self-supervised losses is similar to maximizing the mutual information between a downstream task and node representations, we propose Task-Aware Contrastive Loss with XGBoost Sampler (XTCL). The proposed {\xgs} increases the probability of selecting nodes that enhance the mutual information between node representation and a downstream task as positive examples for contrastive loss. It learns how the downstream task relates to the semantic relations in the graph and actively adjusts the number of positive examples for each semantic relation. The performance results on both node classification and link prediction have shown that XTCL can significantly outperform the state-of-the-art unsupervised and self-supervised models and significantly outperform supervised models in $90\%$ of our experiments. Additionally, the interpretability of XTCL is enhanced by the weights assigned to each semantic relation. These weights indicate the importance of each semantic relation to a downstream task.

\bibliographystyle{ACM-Reference-Format}
\bibliography{sample-base}

\appendix

\section{Experiment Setup}
\label{apx:exp_env}
All codes and scripts are run in the environment with  Intel(R) Xeon(R) Platinum 8358 CPU @ 2.60GHz and NVIDIA A10 GPU. The scripts are available on GitHub\footnote{Github link will be provided after publication to preserve anonymity.}. We refer to the official repository to implement all the baseline methods. We tuned for best parameter settings using grid search and the parameter settings for our models is in our GitHub repo.

\section{Semantic Relations}
\label{sec:semantic_relation}

We describe the set of semantic relations $\gR$ used in our experiments in following sections. We are interested in understanding the relations between two nodes $u\in\gV$ and $v\in\gV$. $N^k_u$ is the $k$-hop neighbors of $u$. Note that we use training data to calculate these similarity values. Below we specify the similarity measure for each semantic relation and commonly used definitions.
\begin{itemize}
    \item Adjacency matrix $A$ is a square matrix describing links between nodes. $A_{u,v}=1$ if $(u,v)\in\E$ else $A_{u,v}=0.$ 
    \item $Y\in[0,1]^{|\gV|\times C}$ is the one-hot encoding of class labels, where $C$ is the number of classes.
    \item $X\in\sR^{|\gV|\times D}$ describes the attributes of all nodes.
\end{itemize}

\vspace{1mm}
\noindent\textbf{Has Link (Link).}~\citep{gcn} 
\begin{equation}
    s_{\text{link}}(u,v)=\widetilde{D}^{-\frac{1}{2}}\widetilde{A}\widetilde{D}^{-\frac{1}{2}}[u,v],
\end{equation}
where $\widetilde{A}=A+I$ is the adjacency matrix $A$ with self-loop and $\widetilde{D}_{uu}=\sum_{v}\widetilde{A}_{uv}$ and $[u,v]$ indicates the $(u,v)$-entry of $\widetilde{D}^{-\frac{1}{2}}\widetilde{A}\widetilde{D}^{-\frac{1}{2}}.$ This similarity is inspired by \citep{gcn} in order to put weight on each link.

\vspace{1mm}
\noindent\textbf{PageRank.}~\citep{pagerank}  We can derive $\pi_u$ of personalized PageRank by solving the following equation
\begin{equation*}
    \pi_u=\alpha P\pi_u + (1-\alpha)e_u,
\end{equation*}
where $P$ is the transition probability, $\alpha\in[0,1]$ and $e_u\in[0,1]^{|\gV|}$ is a one-hot vector containing a one at the position corresponding $u$'s entry and zeros otherwise. The similarity in terms of PageRank is 
\begin{equation}
    s_{\text{PageRank}}(u,v)=\pi_{u,v}.
\end{equation}

\vspace{1mm}
\noindent\textbf{Jaccard Similarity (Jacard Sim.).}~\citep{common_neighbor} Many studies have used this relation to predict links between node pairs. This relation uses the number of 1-hop common neighbors between two nodes to estimate the closeness between them.
\begin{equation}
    s^k_{\text{jaccard}}(u,v)=\frac{|N^k_u\cap N^k_v|}{|N^k_u|\times|N^k_v|}.
\end{equation}
We use $k=1$ in our experiments.

\vspace{1mm}
\noindent\textbf{Topology Similarity (Topology Sim.).}~\citep{n2n} This similarity is similar to Jaccard Similary. Topology similarity represents the mutual information between the neighbors of two nodes. Please refer to \citep{n2n} for mathematical definition of topology similarity.

\vspace{1mm}
\noindent\textbf{Shortest Path Distance (Graph Distance).}~\citep{shortest_path} 
\begin{equation}
    s_{\text{graph-distance}}(u,v)=\frac{\text{diameter}(\gG)-d_{G}(u,v)+1}{\text{diameter}(\gG)},
\end{equation}
where $d_G(u,v)$ is the shortest-path distance between $u$ and $v$. Note that $s_{\text{graph-distance}}(u,v)=\text{diameter}(\gG)+1$ when $u$ and $v$ are disconnected in a graph, and hence $s_{\text{graph-distance}}(u,v)=0$.

\vspace{1mm}
\noindent\textbf{Attribute Similarity. (Attr Sim.)} 
\begin{equation}
    s^k_{\text{attribute similarity}}=\cos(x_u, x_v),
\end{equation}
where $x_u\in\sR^{D}$ is the node attributes of $u$ and $D$ is the dimension.

\vspace{1mm}
\noindent\textbf{Attribute Distribution. (1-Hop Attr. Dist.)}~\citep{lp} 
\begin{equation}
    s^k_{\text{attribute distribution}}=\cos(x^k_u, x^k_v),
\end{equation}
where $x^k_u=Normalize(A^kX)_u$ is the $k$-hop neighbor attribute distribution. We use $k=1$ in our experiments.

\vspace{1mm}
\noindent\textbf{Label Distribution. (2-Hop Label Dist.)}~\citep{lp} 
\begin{equation}
    s^k_{\text{attribute distribution}}=\cos(y^k_u, y^k_v),
\end{equation}
where $y^k_u=Normalize(A^kY)_u$ is the $k$-hop neighbor label distribution. We use $k=2$ in our experiments.

\vspace{1mm}
\noindent\textbf{Attribute Label Distribution. (Attr. Label Dist.)}
\begin{equation}
    s^k_{\text{attr. label distribution}}=\cos(\hat{y}^k_u, \hat{y}^k_v),
\end{equation}
where $\hat{y}^k_u=Normalize(S^k_{\text{attr. dist.}}Y)_u$ is the label distribution of the $k$-hop attribute distribution similarity, $S^k_{\text{attr. dist.}}$. This similarity is especially useful for node classification to identify task-specific positive nodes.

\section{Label Perturbation Process}
\label{apx:label_perturb}
We gradually increase the influence of $\hat{r}$ on the classification task by determining a portion of node labels using $\hat{r}$. Initially, when the level of perturbation is low, $\hat{r}$ is less relevant to the node labels, i.e., $\gI(\mY;S_{\hat{r}})$ is low. We gradually increase the influence of $\hat{r}$ on node labels by setting the new label of node $u$ the same as $v$ if $s_{\hat{r}}(u,v)>\eta_{\hat{r}}$ for a pair of nodes $u$ and $v$. As the number of perturbed pairs increases, $\gI(\mY';S_{\hat{r}})$ increases. By varying the number of perturbed nodes, we can control $\gI(\mY';S_{\hat{r}})$.

To increase the influence of the semantic relation $\hat{r}$ on node labels, the label perturbation must happen in a collective manner, e.g., changing the labels of nodes that have high similarity based on the selected semantic relation $\hat{r}$ together. The key lies in the fact that the graph induced by the similarity matrix of $\hat{r}$ should be homophilic to the perturbed nodes~\citep{wrgat, apann, dgcn} to enhance the dependency between $r$ and node labels.  We select one semantic relation defined in Section~\ref{sec:semantic_relation} as the semantic relation to perturb node labels. Our perturbation process is described in Algorithm~\ref{alg:perturb_label}.

The results in Figure~\ref{fig:motivation_exp} are based on the semi-synthetic data generated from Cora (see Table~\ref{tab:dataset}). We keep the structure of Cora and only determine the labels based on 1-hop Attr. Dist., which is less relevant to $\mY$ (also shown in Figure~\ref{fig:weight}) and to the following loss functions. For each point in Figure~\ref{fig:motivation_exp}, we generate three Cora graphs with different perturbed labels $\mY'$ based on $\hat{r}$, and we create five train/test splits (0.1 for training/0.9 for testing) for each graph with perturbed labels. Given one perturbed graph with one train/test split, we train all models based on the training data and evaluate them with test data. The accuracy (ACC) quantifies whether the predicted label for a test node is the same as its label in the test data. The reported accuracy and standard errors are derived from these $15$ runs. 

\begin{algorithm}[t]
\caption{Label Perturbation Process}\label{alg:perturb_label}
\begin{algorithmic}[1]
\REQUIRE the selected signals $\hat{r}\in\R$, its similarity function $s_{\hat{r}}$, similarity threshold $\{\eta_{\hat{r},u}|u\in\gV\}$, number of perturbation nodes $p$, collective ratio $q$, original labels $\mY$
\ENSURE perturbed node labels $\mY'$
\STATE $\mY'=\mY$
\STATE Initialize stack $U$
\STATE Randomly sample $u\in\gV$ and put it into $\gV'$
\WHILE{$|\gV'| < p$}
    \STATE Pop $u$ from $U$
    \IF{$u\notin \gV'$}
        \FORALL{$\{v|s_{\hat{r}}(u,v)>\eta_{\hat{r},u}\}$}
            \STATE $q'\sim \text{Uniform}(0,1)$
            \IF{$q'<=q$}
                \STATE $y'_v\gets y'_u$
            \ENDIF
        \ENDFOR
        \STATE $\gV' = \gV'\cup u$
        \STATE $U = U\cup \{v|s_{\hat{r}}(u,v)>\eta_{\hat{r},u}\}$
    \ENDIF
\ENDWHILE
\end{algorithmic}
\end{algorithm}

\section{PubMed Performance Discussion}
\label{apx:gcn_error}

\begin{table}[t]
    \centering
    \caption{PubMed Accuracy vs Number of Class Labels.}
    \begin{tabular}{ll}
    \hline
        Num Classes & Accuracy \\ \hline
        3 & 91.7 \\
        4 & 86.7 \\ 
        5 & 85.8 \\
        6 & 83.7 \\ 
        7 & 82.0 \\ \hline
    \end{tabular}
    \label{tab:pubmed}
\end{table}

XTCL(GCN) achieves slightly lower performance compared to GCN in PubMed, and we conjecture that this is due to the smaller number of class label values which enables lower generalization error for GCN.

We conducted additional experiments to test this hypothesis in Section~\ref{sec:task_results} We attempted to decrease the number of classes for training and prediction on Cora and observed how the performance varied with the number of classes. The averages of 5 testing runs are as follows:

In Table~\ref{tab:pubmed}, as the number of classes decreases, the performance of GCN increases accordingly. This is an initial result to test the hypothesis, and a comprehensive investigation is required to confirm it.

\end{document}